\documentclass[11pt,a4paper]{article}
\usepackage[hyperref]{acl2018}
\usepackage[page]{appendix}
\usepackage{placeins}
\usepackage{times}
\usepackage{latexsym}

\usepackage{url}

\usepackage{microtype}

\usepackage{qtree}

\usepackage{float}
\usepackage{amsthm, amsmath}
\usepackage{amssymb}
\usepackage{latexsym}

\usepackage{algpseudocode}
\usepackage{algorithm}

\usepackage{caption}
\usepackage{subcaption}

\usepackage{bussproofs} 
\usepackage{stackengine}

\newcommand{\weight}{\score}
\newcommand{\edgex}{\mbox{\textsc{e}}}
\newcommand{\lbl}{{\mbox{\textsc{lbl}}}}
\newcommand{\mlp}{\ensuremath{\mbox{MLP}}}
\newcommand{\softmax}{{\mbox{softmax}}}
\newcommand{\kg}{K\&G}

\newlength\mylen

\makeatletter
\def\th@definition{%
  \thm@notefont{}
  \normalfont 
}
\makeatother

\theoremstyle{definition} 

\theoremstyle{definition}

\usepackage{amssymb}

\usepackage[pdf,singlefile]{graphviz}

\usepackage{setspace}

\usepackage{bm}
\usepackage{xspace}
\usepackage[utf8]{inputenc}

\newcommand{\tkn}[1]{\texttt{#1}}

\newcommand{\tp}[1]{\ensuremath{\tau(#1)}}   
\newcommand{\type}[1]{\ensuremath{[#1]}}

\newcommand{\app}[1]{\text{\textsc{App}}\ensuremath{_{#1}}}  
\newcommand{\modify}[1]{\text{\textsc{Mod}}\ensuremath{_{#1}}\xspace}
\newcommand{\lignore}{\text{\textsc{ignore}}}

\newcommand{\src}[1]{\text{\textsc{#1}}\xspace}                                  
\newcommand{\obj}[1][]{\text{\src{o}}\ensuremath{_{#1}}}               
\newcommand{\subj}{\text{\src{s}}}													

\newcommand{\G}[1]{\ensuremath{G_{\text{#1}}}} 

 \newcommand{\word}[1]{``#1''}
 \newcommand{\nodelabel}[1]{``#1''}
 \newcommand{\Span}[2]{[#1,#2]}
 \newcommand{\oper}{\text{\textsl{f}}}  

\newcommand{\oneover}[2]{\ensuremath{\displaystyle \begin{array}{@{}c@{}} #1 \\
 		\hline #2  \end{array}}}
\newcommand{\infer}[3]{\oneover{#2}{#3} \;\;{\mbox{#1}}}

\newcommand{\pitem}[5]{(\Span{#1}{#2}, #3, #4):#5}
\newcommand{\ftitem}[4]{(#1,#2,#3):#4}

\newcommand{\score}{\ensuremath{\omega}}
\newcommand{\escore}[2]{\ensuremath{\score(#1 \rightarrow #2)}}
\newcommand{\elscore}[3]{\ensuremath{\score(#1 \mid #2 \rightarrow #3)}}

\newcommand{\ci}[1]{\scalebox{0.8}{\ensuremath{\pm #1}}}

\usepackage{tikz}
\usepackage{tikz-qtree}

\usetikzlibrary{graphs}
\usetikzlibrary{shapes,arrows}
\usetikzlibrary{positioning}
\usetikzlibrary{quotes}

\tikzset{snode/.style={
    ellipse,
    minimum size=6mm,
    very thick,
    draw=black,
    font=\rmfamily}}

\tikzset{ssrc/.style={
    ellipse,
    minimum size=6mm,
    very thick,
    fill=black!10,
    draw=black,
    text=black,
    font=\rmfamily}}

\tikzset{sanno/.style={node distance=-1mm,font=\rmfamily\small}}

\tikzset{sedgel/.style={
    color=black,
    sloped, above,
    font=\rmfamily\small}}

\tikzset{sedge/.style={
    thick,
    >=stealth'
}}

\usepackage{array}

\aclfinalcopy 
\def\aclpaperid{845} 

\title{AMR Dependency Parsing with a Typed Semantic Algebra}

\author{Jonas Groschwitz$\strut^* \strut^\dagger$ \hspace{1em}
 Matthias Lindemann$\strut^*$ \hspace{1em}
 Meaghan Fowlie$\strut^*$ \\
\textbf{Mark Johnson$\strut^\dagger$ \hspace{1em}
 Alexander Koller$\strut^*$}\\
$\strut^*$ Saarland University, Saarbr{\"u}cken, Germany
\quad
$\strut^\dagger$ Macquarie University, Sydney, Australia\\
\url{{jonasg|mlinde|mfowlie|koller}@coli.uni-saarland.de}
\\
\url{mark.johnson@mq.edu.au}
    }

\date{}

\begin{document}
\maketitle


\begin{abstract}
  We present a semantic parser for Abstract Meaning Representations
  which learns to parse strings into tree representations of the
  compositional structure of an AMR graph. This allows us to use
  standard neural techniques for supertagging and dependency tree
  parsing, constrained by a linguistically principled type system. We
  present two approximative decoding algorithms, which achieve
  state-of-the-art accuracy and outperform strong baselines.
\end{abstract}



\section{Introduction} \label{sec:introduction}

Over the past few years, Abstract Meaning Representations (AMRs, \newcite{amBanarescuBCGGHKKPS13}) have become a  popular target representation for semantic parsing. AMRs are graphs which describe the predicate-argument structure of a sentence. Because they are graphs and not trees, they can capture reentrant semantic relations, such as those induced by control verbs and coordination. However, it is technically much more challenging to parse a string into a graph than into a tree. For instance, grammar-based approaches \cite{PengSG15,ArtziLZ15} require the induction of a grammar from the training corpus, which is hard because graphs can be decomposed into smaller pieces in far more ways than trees. Neural sequence-to-sequence models, which do very well on string-to-tree parsing \cite{VinyalsKKPSH14}, can be applied to AMRs but face the challenge that graphs cannot easily be represented as sequences \cite{van2017dealing, van2017neural}.

In this paper, we tackle this challenge by making the compositional
structure of the AMR explicit. As in our previous work, \newcite{graph-algebra-17}, we view an AMR as consisting of atomic
graphs representing the meanings of the individual words, which were
combined compositionally using linguistically motivated
operations for combining a head with its arguments and
modifiers. We represent this structure as terms over the \emph{AM algebra} as defined in \newcite{graph-algebra-17}. This previous work had no parser; here we show that the terms of the AM algebra can
be viewed as dependency trees over the string,
and we train a
dependency parser to map strings into such trees, which
we 
then evaluate into
AMRs in a postprocessing step.
The dependency parser
relies on type information, which encodes the semantic valencies of
the atomic graphs, to guide its decisions.


More specifically, we combine a neural supertagger for identifying the
elementary graphs for the individual words with a neural dependency
model along the lines of
\newcite{kiperwasser16:_simpl_accur_depen_parsin_using} for
identifying the 
operations of the algebra. One key challenge is that
the resulting term of the AM algebra must be semantically
well-typed. This makes the decoding problem NP-complete. We present
two approximation algorithms: one which takes the unlabeled dependency
tree as given, and one which assumes that all dependencies are
projective. 
We evaluate on two data sets, achieving state-of-the-art results on one and near state-of-the-art results on the other (Smatch f-scores of 71.0 and 70.2 respectively). Our approach clearly outperforms strong but non-compositional
baselines. 

\textbf{Plan of the paper.} After reviewing related work in Section~\ref{sec:relwork}, we explain the AM algebra in Section~\ref{sec:am-algebra} and extend it to a dependency view in Section~\ref{sec:indexed-am-algebra}. We explain model training in Section~\ref{sec:training} and decoding in Section~\ref{sec:decode}. Section~\ref{sec:evaluation} evaluates a number of variants of our system.


\section{Related Work} \label{sec:relwork}

Recently, AMR parsing has generated considerable research activity, due to the availability of large-scale annotated data
\cite{amBanarescuBCGGHKKPS13} and two successful SemEval Challenges
\cite{may:2016:SemEval,may-priyadarshi:2017:SemEval}.

Methods from dependency parsing have been shown to be very successful
for AMR parsing. For instance, the JAMR parser
\cite{FlaniganTCDS14,flanigan2016cmu} distinguishes \emph{concept
  identification} (assigning graph fragments to words) from
\emph{relation identification} (adding graph edges which connect these
fragments), and solves the former with a supertagging-style method and
the latter with a graph-based dependency parser.  \newcite{foland2017abstract} use a variant
of this method based on an intricate neural model, yielding state-of-the-art results. We go beyond these
approaches by explicitly modeling the compositional structure of the
AMR, which allows the dependency parser to combine AMRs for the words
using a small set of universal operations, guided by the types
of these AMRs.

Other recent methods directly implement a dependency parser for AMRs,
e.g.\ the transition-based model of \newcite{E17-1051}, or postprocess
the output of a dependency parser by adding missing edges
\cite{DuZSW14,wang15:_trans_algor_amr_parsin}. In contrast
to these, our model makes no strong assumptions on the dependency
parsing algorithm that is used; here we choose that of
\newcite{kiperwasser16:_simpl_accur_depen_parsin_using}.

The commitment of our parser to derive AMRs compositionally mirrors
that of grammar-based AMR parsers \cite{ArtziLZ15,PengSG15}. In
particular, there are parallels between the types we use in the AM
algebra and CCG categories (see Section~\ref{rel-to-ccg} for
details). As a neural system, our parser struggles less with coverage
issues than these, and avoids the complex grammar induction process
these models require.

More generally, our use of semantic types to restrict our parser is reminiscent of \newcite{kwiatkowski2010inducing}, \newcite{krishnamurthy2017neural} and \newcite{zhang2017macro}, and the idea of deriving semantic representations from dependency trees is also present in \newcite{reddy2017universal}.



\section{The AM algebra} \label{sec:am-algebra}

A core idea of this paper is to parse a string into a graph by
instead parsing a string into a dependency-style tree representation
of the graph's compositional structure, represented as terms of the
\emph{Apply-Modify (AM) algebra} \cite{graph-algebra-17}.
 
\begin{figure}
\includegraphics[width=\columnwidth]{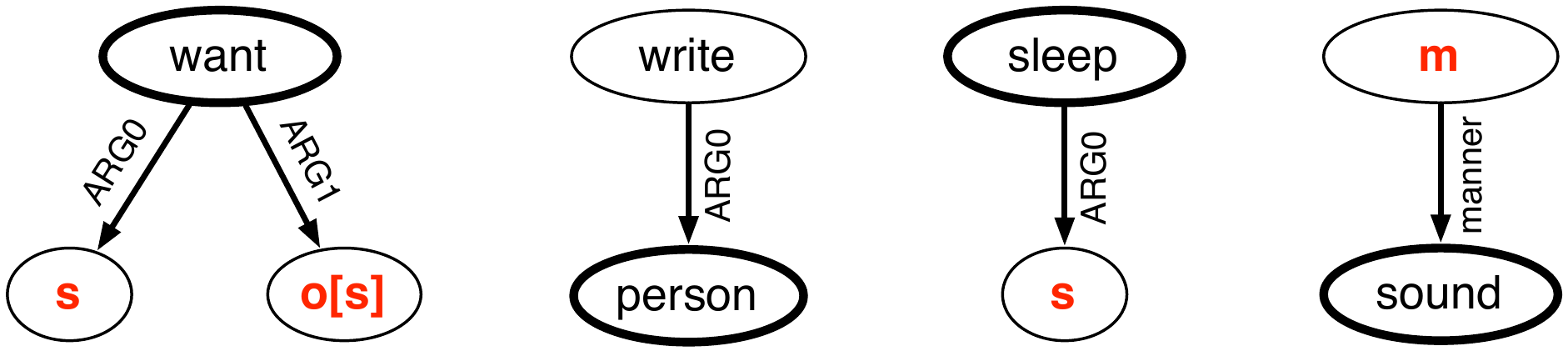}
\caption{Elementary as-graphs \G{want}, \G{writer}, \G{sleep}, and \G{sound} for the words ``want'', ``writer'', ``sleep'', and ``soundly'' respectively.
\label{fig:as-graphs}} 

\vspace{-1em}

\end{figure}

The values of the AM algebra are \textit{annotated s-graphs}, or \emph{as-graphs}: directed graphs
with node and edge labels in which certain nodes have been designated
as \emph{sources} \cite{CourcelleE12} and \emph{annotated} with type
information. Some examples of as-graphs are shown in
Fig.~\ref{fig:as-graphs}. Each as-graph has exactly one \emph{root}, indicated by the bold outline. The sources are
indicated by red labels; for instance, \G{want} has an \src{s}-source
and an \src{o}-source. The annotations, written in square brackets
behind the red source names, will be explained below. We use these sources to mark open argument slots; for example, \G{sleep} in Fig.~\ref{fig:as-graphs} represents an intransitive verb, missing its subject, which will be added at the \src{s}-source.

\begin{figure*}
\includegraphics[width=\textwidth]{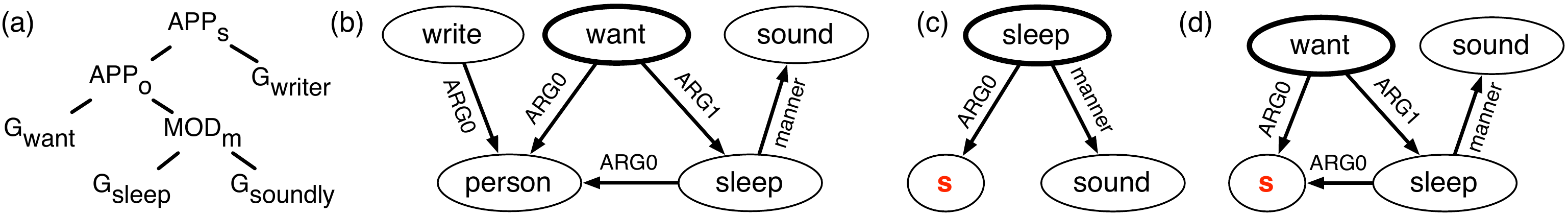}
\caption{(a) An AM-term with its value (b), along with the values for its subexpressions (c) $t_1 = \modify{m}(\G{sleep}, \G{sound})$ and (d) $t_2 = \app{o}(\G{want}, t_1)$. \label{fig:apply-modify} \label{fig:am-term}}

\vspace{-1em}

\end{figure*}

The AM algebra can combine as-graphs with each other using two
linguistically motivated operations: \emph{apply} and \emph{modify}. Apply (\app{}) adds an argument to a predicate. For example, we can add a subject -- the graph \G{writer} in Fig.~\ref{fig:as-graphs} -- to the graph \G{VP }in Fig.~\ref{fig:apply-modify}d using \app{\subj}, yielding the complete AMR in Fig.~\ref{fig:apply-modify}b. Linguistically, this is like filling the subject (\subj) slot of the predicate \textit{wants to sleep soundly} with the argument \textit{the writer}.
In general, for a source $a$, $\app{a}(G_P, G_A)$, combines the as-graph $G_P$
representing a predicate, or head, with the as-graph $G_A$, which represents an argument. It does this by plugging the root node of $G_A$ into the
$a$-source $u$ of $G_P$ -- that is, the node $u$ of $G_P$ marked with source $a$. The root of the resulting as-graph $G$
is the root of $G_P$, and we remove the $a$ marking on $u$, since that slot is now filled.

The modify operation (\modify{}) adds a modifier to a graph. For example, we can combine two elementary graphs from Fig.~\ref{fig:as-graphs} with \modify{m}(\G{sleep}, \G{sound}), yielding the graph in Fig.~\ref{fig:apply-modify}c. The \src{m}-source of the modifier \G{soundly} attaches to the root of \G{sleep}. The root of the result 
is the same as the root of \G{sleep} in the same sense that a verb phrase with an adverb modifier is still a verb phrase. 
In general, $\modify{a}(G_H, G_M)$, combines a head $G_H$ with a modifier $G_M$.
It plugs the root of $G_H$ into the $a$-source $u$ of $G_M$. Although
this may add incoming edges to the root of $G_H$, that node is still
the root of the resulting graph $G$. We remove the $a$ marking from $G_M$.

In both \app{} and \modify{}, if there is any other source $b$ which is present in both graphs, the nodes marked with $b$ are unified with each other. For example, when \G{want} is
\src{o}-applied to $t_1$ in Fig.~\ref{fig:apply-modify}d, the
\src{s}-sources of the graphs for ``want'' and ``sleep soundly'' are
unified into a single node, creating a reentrancy. This falls out of the definition of \textit{merge} for s-graphs which formally underlies both operations (see \cite{CourcelleE12}).

Finally, the AM algebra uses \emph{types} to restrict its operations.
Here we define the \emph{type} of an as-graph as the set of its sources with their annotations\footnote{See \cite{graph-algebra-17} for a more formally complete definition.}; thus for
example, in Fig.~\ref{fig:as-graphs}, the graph for ``writer'' has the empty 
type \type{\;}, \G{sleep} has type \type{\subj}, and \G{want} has type
\type{\subj,\obj{[\subj]}}. Each source in an as-graph specifies with its \textit{annotation} the type of the
as-graph which is plugged into it via \app{}. In other words, for a source $a$, we may only $a$-apply $G_P$ with $G_A$
if the annotation of the $a$-source in $G_P$ matches the type of
$G_A$. For example, the \src{o}-source
of \G{wants} (Fig.~\ref{fig:as-graphs}) requires that we plug in an as-graph of type
$\type{\src{s}}$; observe that this means that the reentrancy in Fig.~\ref{fig:apply-modify}b is lexically
specified by the control verb ``want''. All other source nodes in Fig.~\ref{fig:as-graphs} have no annotation,
indicating a type requirement of \type{\;}. 

Linguistically, modification is optional; we therefore want the modified graph to be derivationally just like the unmodified graph, in that exactly the same operations can apply to it. In a typed algebra, this means \modify{} should not change the type of the head. \modify{a} therefore requires that the modifier $G_M$ have no sources not already present in the head $G_H$, except $a$, which will be deleted anyway.

As in any algebra, we can build \emph{terms} from constants (denoting elementary
as-graphs) by recursively combining them with the operations of the AM
algebra. By evaluating the operations bottom-up, we obtain an as-graph as the value of such a term; see Fig.~\ref{fig:am-term} for an example. However, as discussed above, an operation in the term may be undefined due to a type mismatch. We call an AM-term
\emph{well-typed} if all its operations are defined. Every well-typed AM-term evaluates to an as-graph. Since the
applicability of an AM operation depends only on the types, we also
write $\tau = \oper(\tau_1, \tau_2)$ if as-graphs of type $\tau_1$ and
$\tau_2$ can be combined with the operation \oper{} and the result has
type $\tau$.


\newcommand{\nt}[1]{\mathrm{#1}}
\newcommand{\sback}{\backslash}
\newcommand{\sforw}{/}

\paragraph{Relationship to CCG.} \label{rel-to-ccg}%
There is close relationship between the types of the AM algebra and the categories of CCG. A type $\type{\src{s},\src{o}}$ specifies that the as-graph needs to be applied to two arguments to be semantically complete, similar a CCG category such as $\nt{S} \sback \nt{NP} \sforw \nt{NP}$, where a string needs to be applied to two NP arguments to be syntactically complete.
However, AM types govern the combination of \emph{graphs}, while CCG categories control the combination of strings. This relieves AM types of the need to talk about word order; there are no ``forward'' or ``backward'' slashes in AM types, and a smaller set of operations. Also, the AM algebra spells out raising and control phenomena more explicitly in the types.

\section{Indexed AM terms}
\label{sec:indexed-am-algebra}

\begin{figure}
\vspace{-2ex}
\includegraphics[width=\columnwidth]{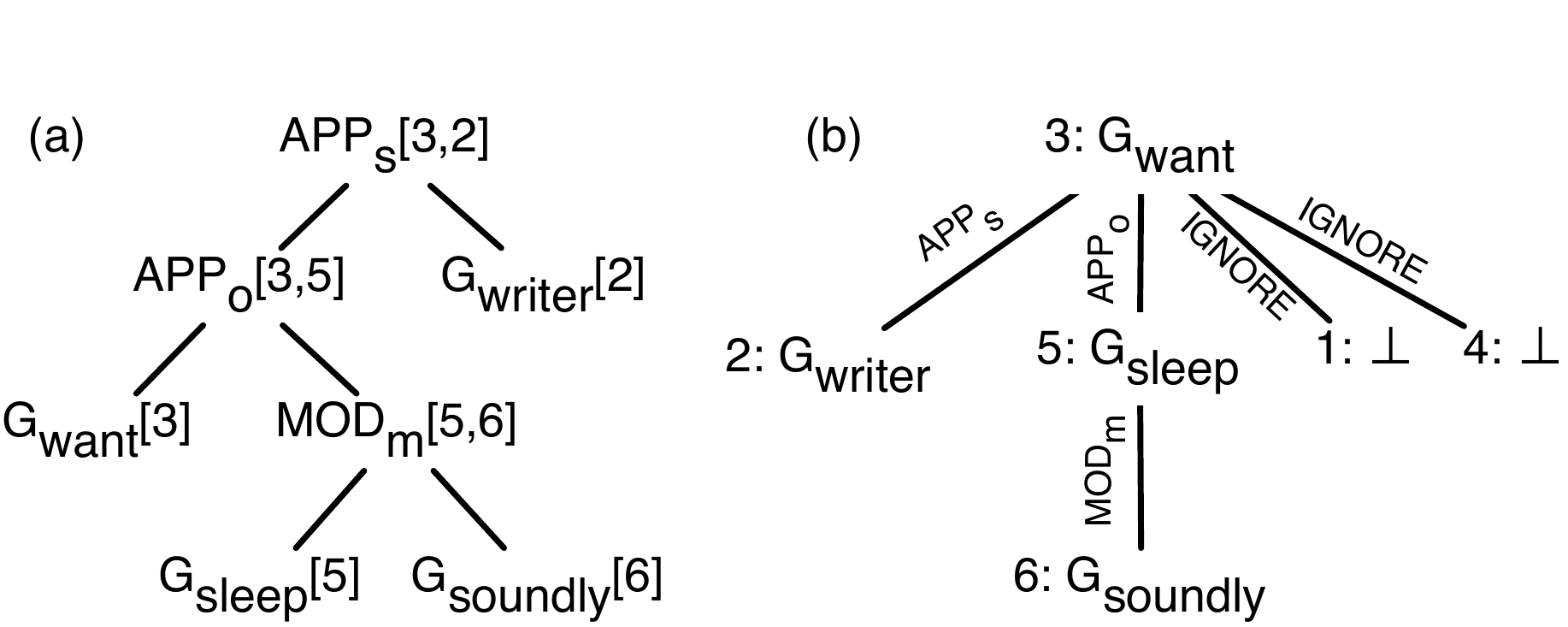}
\caption{(a) An indexed AM term and (b) an AM dependency tree, linking the term in Fig.~\ref{fig:am-term};a to the sentence \word{The writer wants to sleep soundly}. \label{fig:indexed-am-term}}

\vspace{-1em}

\end{figure}

In this paper, we connect AM terms to the input string $w$ for which
we want to produce a graph. We do this in an \emph{indexed AM term}, exemplified in Fig.~\ref{fig:indexed-am-term}a. We assume that every elementary as-graph $G$ at a leaf
represents the meaning of an individual word token $w_i$ in $w$,
and write $G[i]$ to
annotate the leaf $G$ with the index $i$ of this token. This induces
a connection between the nodes of the AMR and the tokens of the
string, in that the label of each node was contributed by the
elementary as-graph of exactly one token.

We define the \emph{head index} of a subtree $t$ to be the index of
the token which contributed the root of the as-graph to which $t$
evaluates. For a leaf with annotation $i$, the head index is $i$; for
an \app{} or \modify{} node, the head index is the head index of the left
child, i.e.\ of the head argument. We annotate each \app{} and \modify{}
operation with the head index of the left and right subtree.

\subsection{AM dependency trees} \label{sec:am-dependency-trees}

We can represent indexed AM terms more compactly as \emph{AM
  dependency trees}, as shown in Fig.~\ref{fig:indexed-am-term}b. The
nodes of such a dependency tree are the tokens of $w$. We draw an edge
with label \oper{} from $i$ to $k$ if there is a node with label $\oper[i,k]$
in the indexed AM term. For example, the tree in \ref{fig:indexed-am-term}b has an edge labeled \modify{m}  from 5 (\G{sleep}) to 6 (\G{soundly}) because there is a node in the term in \ref{fig:indexed-am-term}a labeled $\modify{m}[5,6]$. The same AM dependency tree may represent multiple indexed AM terms, because the order of apply and modify operations is not specified in the dependency tree. However, it can be shown that all well-typed AM terms that map to the same AM dependency tree evaluate to the same as-graph. We define a \emph{well-typed} AM dependency tree as one that represents a well-typed AM term.

Because not all words in the sentence contribute to the AMR, we  include a mechanism for ignoring words in the input. As a special case, we allow the constant $\bot$, which represents a dummy as-graph (of type $\bot$) which we use as the semantic value of words without a semantic value in the AMR. We furthermore allow the edge label \lignore\ in an AM dependency tree, where $\lignore(\tau_1,\tau_2) = \tau_1$ if $\tau_2 = \bot$ and is undefined otherwise; in particular, an AM dependency tree with \lignore\ edges is only well-typed if all \lignore\ edges point into $\bot$ nodes. We keep all other operations $\oper(\tau_1,\tau_2)$ as is, i.e. they are undefined if either $\tau_1$ or $\tau_2$ is $\bot$, and never yield $\bot$ as a result.  When reconstructing an AM term from the AM dependency tree, we skip \lignore\ edges, such that the subtree below them will not contribute to the overall AMR.

\subsection{Converting AMRs to AM terms}
\label{sec:converting-amrs-am}

In order to train a model that parses sentences into AM dependency
trees, we need to convert an AMR corpus -- in which sentences are
annotated with AMRs -- into a treebank of AM dependency trees. We do
this in three steps: first, we break each AMR up into elementary graphs
and identify their roots; second, we assign sources and annotations to
make elementary as-graphs out of them; and third, combine them into indexed
AM terms.

For the first step, an aligner uses hand-written heuristics to
identify the string token to which each node in the AMR corresponds
(see Section C in the Supplementary Materials for details). We proceed
in a similar fashion as the JAMR aligner \cite{FlaniganTCDS14}, i.e.\
by starting from high-confidence token-node pairs and then extending
them until the whole AMR is covered. Unlike the JAMR aligner, our
heuristics ensure that exactly one node in each elementary graph is
marked as the root, i.e.\ as the node where other graphs can attach
their edges through \app{} and \modify{}. When an edge connects 
nodes of two different elementary graphs, we use the ``blob
decomposition'' algorithm of \newcite{graph-algebra-17} to decide to
which elementary graph it belongs. For the example AMR in
Fig.~\ref{fig:apply-modify}b, we would obtain the graphs in
Fig.~\ref{fig:as-graphs} (without source annotations). Note that ARG
edges belong with the nodes at which they start, whereas the
``manner'' edge in \G{soundly} goes with its target.

In the second step we assign source names and annotations to the unlabeled
nodes of each elementary graph. Note that the annotations are crucial
to our system's ability to generate graphs with reentrancies. We
mostly follow the algorithm of \newcite{graph-algebra-17}, which determines necessary annotations based on the structure of the given graph. The algorithm chooses
each source name depending on the incoming edge label. For instance,
the two leaves of \G{want} can have the source labels \src{s} and
\src{o} because they have incoming edges labeled ARG0 and
ARG1. However, the Groschwitz algorithm is not deterministic: It
allows object promotion (the sources for an ARG3 edge may be \src{o3},
\src{o2}, or \src{o}), unaccusative subjects (promoting the minimal object to \subj{} if the elementary graph
contains an ARGi-edge ($i>0$) but no ARG0-edge \cite{perlmutter1978impersonal}), and passive alternation (swapping \src{o} and
\src{s}). To make our as-graphs more consistent, we prefer constants that promote objects
as far as possible, use unaccusative subjects, and no passive alternation, but still allow constants that do not satisfy these conditions if necessary. This increased our Smatch score significantly.

Finally, we choose an arbitrary AM dependency tree that combines the chosen
elementary as-graphs into the annotated AMR; in practice, the
differences between the trees seem to be negligible.\footnote{Indeed, we conjecture that for a fixed set of constants and a fixed AMR, there is only one dependency tree.}



\section{Training} \label{sec:training}


We can now model the AMR parsing task as the problem of computing the
best well-typed AM dependency tree $t$ for a given sentence
$w$. Because $t$ is well-typed, it can be decoded into an (indexed) AM
term and thence evaluated to an as-graph.

We describe $t$ in terms of the elementary as-graphs $G[i]$ it uses
for each token $i$ and of its edges $\oper[i,k]$. We assume a
node-factored, edge-factored model for the \emph{score} $\score(t)$ of
$t$:
\begin{equation} \label{eq:score}
\score(t) = \sum_{1 \leq i \leq n} \score(G[i]) + \sum_{\oper[i,k] \in E} \score(\oper[i,k]),
\end{equation}
\noindent
where the edge weight further decomposes into the sum $\score(\oper[i,k]) =
\escore{i}{k} + \elscore{\oper}{i}{k}$ of a score $\escore{i}{k}$ for the
presence of an edge from $i$ to $k$ and a score $\elscore{\oper}{i}{k}$
for this edge having label \oper. Our aim is to compute the well-typed
$t$ with the highest score.

\begin{figure}
\centering
\includegraphics[width=0.9\columnwidth]{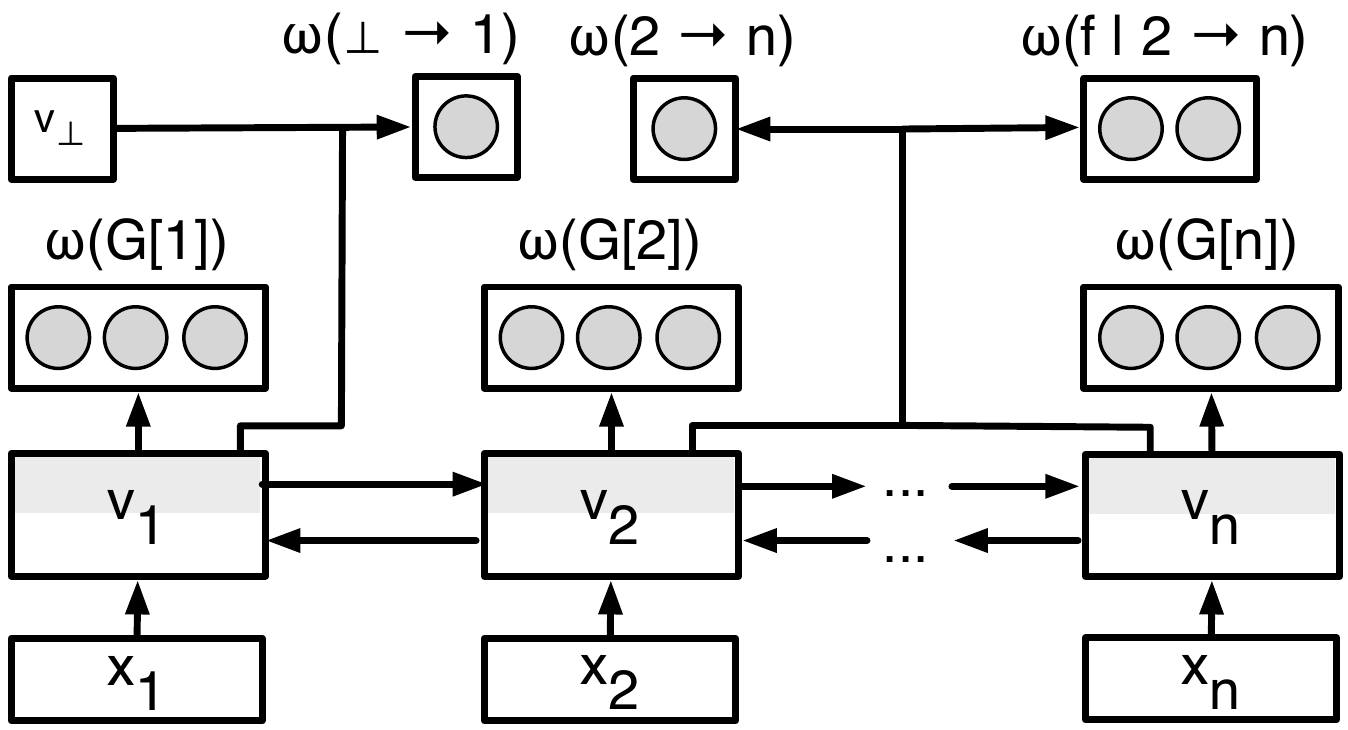}
\caption{Architecture of the neural taggers.} \label{fig:nn}

\vspace{-1em}

\end{figure}

We present three models for $\score$: one for the graph scores and two
for the edge scores. All of these are based on a two-layer
bidirectional LSTM, which reads inputs $\mathbf{x} = (x_1,\ldots,x_n)$
token by token, concatenating the hidden states of the forward and the
backward LSTMs in each layer. On the second layer, we thus obtain
vector representations
$v_i = \mbox{BiLSTM}(\mathbf{x}, i)$
for the individual input tokens (see Fig.~\ref{fig:nn}). 
Our models differ in the inputs $\mathbf{x}$ and the way they predict scores from the $v_i$.

\subsection{Supertagging for elementary as-graphs}
\label{sec:supertagging}

We construe the prediction of the as-graphs $G[i]$ for each input
position $i$ as a supertagging task \cite{lewis2016lstm}. The
supertagger reads inputs $x_i = (w_i, p_i, c_i)$, where $w_i$ is the
word token, $p_i$ its POS tag, and $c_i$ is a character-based LSTM
encoding of $w_i$. We use pretrained GloVe embeddings \cite{pennington2014glove}
concatenated with learned embeddings for $w_i$,
and learned embeddings for $p_i$.

To predict the score for each elementary as-graph out of a set of $K$
options, we add a $K$-dimensional output layer as follows:
$$\weight(G[i]) = \log \softmax(W \cdot v_i + b)$$
and train the neural network using a cross-entropy loss function. This
maximizes the likelihood of the elementary as-graphs in the training
data.

\subsection{Kiperwasser \& Goldberg edge model}
\label{sec:kiperw-goldb-model} \label{sec:pure-depend-model}

Predicting the edge scores amounts to a dependency parsing problem. We
chose the dependency parser of
\newcite{kiperwasser16:_simpl_accur_depen_parsin_using}, henceforth
\kg, to learn them, because of its accuracy and its fit with our
overall architecture. The \kg\ parser scores the potential edge from
$i$ to $k$ and its label from the concatenations of $v_i$ and $v_k$:
$$
\begin{array}{rcl}
  \mlp_\theta(v) &= &W_2 \cdot \tanh(W_1 \cdot v + b_1) + b_2\\
  \weight(i \rightarrow k) &= &\mlp_{\edgex} (v_i \circ v_k)\\
  \weight(\oper{} \mid i \rightarrow k) &= &\mlp_\lbl (v_i \circ v_k)
\end{array}
$$
We use inputs $x_i = (w_i, p_i, \tau_i)$ including the type $\tau_i$
of the supertag $G[i]$ at position $i$, using trained embeddings for all three. 
At evaluation time, we use the best scoring supertag according to the model of Section~\ref{sec:supertagging}. At training time, we sample from $q$, where $q(\tau_i) = (1-\delta) + \delta \cdot p(\tau_i|p_i, p_{i-1})$, $q(\tau) = \delta \cdot p(\tau |p_i, p_{i-1})$ for any $\tau \neq \tau_i$ and $\delta$ is a hyperparameter controlling the bias towards the aligned supertag.
We train the model using \kg
's original DyNet implementation. Their algorithm uses a hinge loss
function, which maximizes the score difference between the
gold dependency tree and the best predicted dependency tree, and
therefore requires parsing each training instance in each iteration.
Because the AM dependency trees are highly non-projective, we replaced
the projective parser used in the off-the-shelf implementation by the
Chu-Liu-Edmonds algorithm implemented in the TurboParser
\cite{martins2010turbo}, improving the LAS on the development set by
30 points.

\subsection{Local edge model}
\label{sec:simpl-depend-model}

We also trained a \emph{local} edge score model, which uses a
cross-entropy rather than a hinge loss and therefore avoids the
repeated parsing at training time. Instead, we follow the intuition
that every node in a dependency tree has at most one incoming edge,
and train the model to score the correct incoming edge as high as
possible. This model takes inputs $x_i = (w_i,p_i)$.

We define the edge and edge label scores as in
Section~\ref{sec:kiperw-goldb-model}, with tanh replaced by ReLU.  We
further add a learned parameter $v_\bot$ for the ``LSTM
embedding'' of a nonexistent node, obtaining scores
$\score(\bot \rightarrow k)$ for $k$ having no incoming edge.

To train $\escore{i}{k}$, we collect all scores for edges ending at the same node $k$
into a vector $\escore{\bullet}{k}$. We then minimize the
cross-entropy loss for the gold edge into $k$ under
$\softmax(\escore{\bullet}{k})$, maximizing the likelihood of the gold
edges. To train the labels $\elscore{\oper}{i}{k}$, we simply minimize the cross-entropy loss of the actual edge labels \oper{} of the edges which
are present in the gold AM dependency trees.

The PyTorch code for this and the supertagger are available at \url{bitbucket.org/tclup/amr-dependency}.

\section{Decoding}\label{sec:decode}

Given learned estimates for the graph and edge scores, we now tackle the challenge of computing the best well-typed dependency tree $t$ for the input string $w$, under the score model (equation \eqref{eq:score}). The requirement that $t$ must be well-typed is crucial to ensure that it can be evaluated to an AMR graph, but as we show in the Supplementary Materials (Section A), makes the decoding problem NP-complete. Thus, an exact algorithm is not practical. In this section, we develop two different approximation algorithms for AM dependency parsing: one which assumes the (unlabeled) dependency tree structure as known, and one which assumes that the AM dependency tree is projective.

\begin{figure}
  \centering {\small
  \infer{Init}{
    s = \score(G[i])
    \quad
    G \neq \bot
  }{
    \pitem{i}{i+1}{i}{\tp{G}}{s}
  }\strut\\\strut

  \infer{Skip-R}{
    \pitem{i}{k}{r}{\tau}{s}
    \quad
    s' = \score(\bot[k])
  }{
    \pitem{i}{k+1}{r}{\tau}{s+s'}
  }\strut\\\strut

  \infer{Skip-L}{
    \pitem{i}{k}{r}{\tau}{s}
    \quad
    s' = \score(\bot[i-1])
  }{
    \pitem{i-1}{k}{r}{\tau}{s+s'}
  }

  \begin{prooftree}
  \AxiomC{\stackanchor{$\pitem{i}{j}{r_1}{\tau_1}{s_1} \quad \pitem{j}{k}{r_2}{\tau_2}{s_2}$}
                {$\tau = \oper(\tau_1, \tau_2) \;\mbox{defined} \quad s = \score(\oper[r_1, r_2])$}
        }
  
  \RightLabel{Arc-R [\oper]}
  \UnaryInfC{$\pitem{i}{k}{r_1}{\tau}{s_1+s_2+s}$}
  \end{prooftree}

  
  \begin{prooftree}
  \AxiomC{\stackanchor{$\pitem{i}{j}{r_1}{\tau_1}{s_1} \quad \pitem{j}{k}{r_2}{\tau_2}{s_2}$}
                {$\tau = \oper(\tau_2, \tau_1) \;\mbox{defined} \quad s = \score(\oper[r_2, r_1])$}
        }
  
  \RightLabel{Arc-L [\oper]}
  \UnaryInfC{$\pitem{i}{k}{r_2}{\tau}{s_1+s_2+s}$}
  \end{prooftree}

  
  }
  \caption{Rules for the projective decoder.}
  \label{fig:rules-projective}
 
 \vspace{-1em}
  
\end{figure}


\subsection{Projective decoder} \label{sec:decoder-projective}

The projective decoder assumes that the AM dependency tree is
projective, i.e.\ has no crossing dependency edges. Because of this
assumption, it can recursively combine adjacent substrings using
dynamic programming. The algorithm is shown in
Fig.~\ref{fig:rules-projective} as a parsing schema
\cite{shieber95:_princ_implem_deduc_parsin}, which derives items of
the form $(\Span{i}{k}, r, \tau)$ with scores $s$. An item
represents a well-typed derivation of the substring from $i$ to $k$
with head index $r$, and which evaluates to an as-graph of type
$\tau$. 

The parsing schema consists of three types of rules. First, the Init
rule generates an item for each graph fragment $G[i]$ that the
supertagger predicted for the token $w_i$, along with the score and
type of that graph fragment. Second, given items for adjacent
substrings $\Span{i}{j}$ and $\Span{j}{k}$, the Arc rules apply an
operation \oper{} to combine the indexed AM terms for the two substrings,
with Arc-R making the left-hand substring the head and the right-hand
substring the argument or modifier, and Arc-L the other way around. We
ensure that the result is well-typed by requiring that the types can
be combined with \oper{}. Finally, the Skip rules allow us to extend a
substring such that it covers tokens which do not correspond to a
graph fragment (i.e., their AM term is $\bot$), introducing \lignore\
edges. After all possible items have been derived, we extract the best
well-typed tree from the item of the form $(\Span{1}{n}, r, \tau)$
with the highest score, where $\tau = \type{\;}$.

Because we keep track of the head indices, the projective decoder is a
bilexical parsing algorithm, and shares a parsing complexity of
$O(n^5)$ with other bilexical algorithms such as the Collins
parser. It could be improved to a complexity of $O(n^4)$ using the
algorithm of \newcite{eisner99:_effic}.

\subsection{Fixed-tree decoder} \label{sec:fixed-tree-decoder}

The fixed-tree decoder computes the best unlabeled dependency
tree $t_r$ for $w$, using the edge scores \escore{i}{k}, and then
computes the best AM dependency tree for $w$ whose unlabeled version is
$t_r$. The Chu-Liu-Edmonds algorithm produces a forest of
dependency trees, which we want to combine into $t_r$. 
We choose the tree whose root $r$ has the highest score for being the root of the AM dependency tree and make the roots of all others children of $r$. 

\begin{figure}
  \centering {\small
  
  \infer{Init}{
    s = \weight(G[i])
  }{
    \ftitem{i}{\emptyset}{\tau(G)}{s}
  }
  \begin{prooftree}
    \AxiomC{
        \stackanchor{
                \stackanchor{ $\ftitem{i}{C_1}{\tau_1}{s_1} \quad \ftitem{k}{Ch(k)}{\tau_2}{s_2}$}
                  {$k \in Ch(i) \backslash C_1$}}
          {$\tau = \oper(\tau_1, \tau_2) \;\mbox{defined} \quad s = \score(\oper[i,k])$}}

    \RightLabel{Edge[\oper]}
    \UnaryInfC{$\ftitem{i}{C_1 \cup \{k\}}{\tau}{s_1+s_2+s}$}
\end{prooftree}

}
  \caption{Rules for the fixed-tree decoder.}
  \label{fig:rules-fixed-tree}
  
  \vspace{-1em}
  
\end{figure}


At this point, the shape of $t_r$ is fixed. We choose supertags for the
nodes and edge labels for the edges by traversing $t_r$ bottom-up,
computing types for the subtrees as we go along. Formally, we apply
the parsing schema in Fig.~\ref{fig:rules-fixed-tree}. It uses items
of the form $\ftitem{i}{C}{\tau}{s}$, where $1 \leq i \leq n$ is a
node of $t_r$, $C$ is the set of children of $i$ for which we have
already chosen edge labels, and $\tau$ is a type. 
We write $Ch(i)$ for the
set of children of $i$ in $t_r$.

The Init rule generates an item for each graph that the supertagger
can assign to each token $i$ in $w$, ensuring that every token is also
assigned $\bot$ as a possible supertag. The Edge rule labels an edge
from a parent node $i$ in $t_r$ to one of its children $k$, whose
children already have edge labels. As above, this rule ensures that a
well-typed AM dependency tree is generated by locally checking the
types. In particular, if all types $\tau_2$ that can be derived for
$k$ are incompatible with $\tau_1$, we fall back to an item for $k$
with $\tau_2 = \bot$ (which always exists), along with
an \lignore\ edge from $i$ to $k$.

The complexity of this algorithm is
$O(n \cdot 2^d \cdot d)$, where $d$ is the maximal arity of the nodes in $t_r$.


\section{Evaluation} \label{sec:evaluation}

We evaluate our models on the LDC2015E86 and LDC2017T10\footnote{\url{https://catalog.ldc.upenn.edu/LDC2017T10}, identical to LDC2016E25.} datasets (henceforth ``2015" and ``2017"). Technical details and hyperparameters of our implementation  can
be found in Sections B to D of the Supplementary Materials.

\subsection{Training data}

The original LDC datasets pair strings with AMRs. We convert
each AMR in the training and development set into an AM
dependency tree, using the procedure of
Section~\ref{sec:converting-amrs-am}. About 10\% of the training
instances cannot be split into elementary as-graphs by our aligner; we
removed these from the training data. Of the remaining AM dependency
trees, 37\% are non-projective.

Furthermore, the AM algebra is designed to handle short-range
reentrancies, modeling grammatical phenomena such as control and
coordination, as in the derivation in Fig.~\ref{fig:am-term}. It
cannot easily handle the long-range reentrancies in AMRs which are
caused by coreference, a non-compositional phenomenon.\footnote{As \newcite{E17-1051} comment: ``A valid criticism of AMR is that these two reentrancies are of a completely different type, and should not be collapsed together.''} We remove such
reentrancies from our training data (about 60\% of the roughly 20,000
reentrant edges). Despite this, our model performs well on reentrant edges (see Table~\ref{table:damonte}).

%

\subsection{Pre- and postprocessing}\label{sec:post}


We use simple pre- and postprocessing steps to handle rare words and some AMR-specific patterns. In AMRs, named entities follow a pattern shown in Fig.~\ref{fig:namedEnt}. Here the named entity is of type \nodelabel{person}, has a name edge to a \nodelabel{name} node whose children spell out the tokens of ``Agatha Christie'', and a link to a wiki entry. Before training, we replace each \nodelabel{name} node, its children, and the corresponding span in the sentence
with a special \tkn{NAME} token, and we completely remove wiki edges. In this example, this leaves us with only a \nodelabel{person} and a \tkn{NAME} node. Further, we replace numbers and some date patterns with \tkn{NUMBER} and \tkn{DATE} tokens.
On the training
data this is straightforward, since names and dates are explicitly annotated in the AMR. At
evaluation time, we detect dates and numbers with regular expressions,
and names with Stanford CoreNLP \cite{ManningSBFBM14}. We also use
Stanford CoreNLP for our POS tags.

Each elementary as-graph generated by the procedure of
Section~\ref{sec:converting-amrs-am} has a unique node whose label
corresponds most closely to the aligned word (e.g.\ the
\nodelabel{want} node in \G{want} and the \nodelabel{write} node in
\G{writer}). We replace these node labels with \tkn{LEX} in
preprocessing, reducing the number of different elementary as-graphs
from 28730 to 2370. We factor the supertagger model of
Section~\ref{sec:supertagging} such that the unlexicalized version of
$G[i]$ and the label for \tkn{LEX} are predicted
separately.

At evaluation, we re-lexicalize all \tkn{LEX} nodes in the predicted AMR. For words that
were frequent in the training data (at least 10 times), we take the
supertagger's prediction for the label. For rarer words, we use simple
heuristics, explained in the Supplementary Materials (Section
D). For names, we just look up name nodes with their children and wiki entries
observed for the name string in the training data,
and for unseen names use
the literal tokens as the name, and no wiki entry. Similarly, we collect the type for each encountered name (e.g. \nodelabel{person} for ``Agatha Christie''), and correct it in the output if the tagger made a different prediction.
We recover dates and
numbers straightforwardly.

\subsection{Supertagger accuracy}

All of our models rely on the supertagger to predict elementary
as-graphs; they differ only in the edge scores. We evaluated the
accuracy of the supertagger on the converted development set (in which
each token has a supertag) of the 2015 data set, and achieved an accuracy of 73\%. The
correct supertag is within the supertagger's 4 best predictions for
90\% of the tokens, and within the 10 best for 95\%.

Interestingly, supertags that introduce grammatical reentrancies are
predicted quite reliably, although they are relatively rare in the
training data. The elementary as-graph for subject control verbs (see
\G{want} in Fig.~\ref{fig:as-graphs}) accounts for only 0.8\% of
supertags in the training data, yet 58\% of its occurrences in the
development data are predicted correctly (84\% in 4-best). The
supertag for VP coordination (with type \type{\src{op1}[\src{s}], \src{op2}[\src{s}]}) makes up for
0.4\% of the training data, but 74\% of its occurrences are recognized
correctly (92\% in 4-best). Thus the prediction of informative types
for individual words is feasible.

\subsection{Comparison to Baselines}

\textbf{Type-unaware fixed-tree baseline.} The fixed-tree decoder is built to ensure
well-typedness of the predicted AM dependency trees. To investigate to what
extent this is required, we consider a baseline which just adds the
individually highest-scoring supertags and edge labels to the
unlabeled dependency tree $t_u$, ignoring types. This leads to AM dependency trees
which are not well-typed for 75\% of the sentences (we fall back to the largest well-typed subtree in these
cases). Thus, an off-the-shelf dependency parser can reliably predict
the tree structure of the AM dependency tree, but correct supertag and
edge label assignment requires a decoder which takes the types into
account.

\textbf{JAMR-style baseline.} Our elementary as-graphs differ from the
elementary graphs used in JAMR-style algorithms in that they contain
explicit source nodes, which restrict the way in which they can be
combined with other as-graphs. We investigate the impact of this
choice by implementing a strong JAMR-style baseline. We adapt the
AMR-to-dependency conversion of Section~\ref{sec:converting-amrs-am}
by removing all unlabeled nodes with source names from the elementary
graphs. For instance, the graph \G{want} in Fig.~\ref{fig:as-graphs}
now only consists of a single \nodelabel{want} node. We then
aim to directly predict AMR edges between these graphs, using a
variant of the local edge scoring model of
Section~\ref{sec:simpl-depend-model} which learns scores for each edge in
isolation. (The assumption for the original local
model, that each node has
only one incoming edge, does not apply here.) 

When parsing a string, we choose the highest-scoring supertag for each
word; there are only 628 different supertags in this setting, and
1-best supertagging accuracy is high at 88\%. We then follow the JAMR
parsing algorithm by predicting all edges whose score is over a
threshold (we found -0.02 to be optimal) and then adding edges until
the graph is connected. Because we do not predict which node is the
root of the AMR, we evaluated this model as if it always predicted the
root correctly, overestimating its score slightly.

 \begin{table}
        \centering
        {\small
           \begin{tabular}{|l|r|r|}
            \hline
             \textbf{Model} & \multicolumn{1}{c|}{\textbf{2015}} & \multicolumn{1}{c|}{\textbf{2017}}\\
            \hline
            Ours &&\\
            \hline
             ~local edge + projective decoder & $70.2 \ci{0.3}$ & $\textbf{71.0} \ci{0.5}$\\
             ~local edge + fixed-tree decoder & $69.4 \ci{0.6}$ & $70.2 \ci{0.5}$\\ 
             ~\kg\  edge + projective decoder & $68.6 \ci{0.7}$ & $69.4 \ci{0.4}$ \\
             ~\kg\  edge + fixed-tree decoder  & $69.6 \ci{0.4}$ & $69.9 \ci{0.2}$ \\
             
            \hline
            Baselines &&\\
            \hline
             ~fixed-tree (type-unaware) &  $26.0 \ci{0.6}$ & $27.9 \ci{0.6}$ \\
             ~JAMR-style & $66.1$ & 66.2 \\
             \hline
             Previous work &&\\
            \hline
    
             ~CAMR \cite{wang15:_trans_algor_amr_parsin} & 66.5 & -\\
             ~JAMR \cite{flanigan2016cmu} & 67 & -\\
             ~\newcite{E17-1051} & 64 & -\\
             ~\newcite{van2017neural} & 68.5 & \textbf{71.0}\\
             ~\newcite{foland2017abstract} & \textbf{70.7} & - \\
             ~\newcite{buys2017oxford} & - & 61.9\\
            \hline 
    \end{tabular}
      }
        \caption{2015 \&{} 2017 test set Smatch scores} \label{table:smatch}

\vspace{-1em}

\end{table}

\subsection{Results}\label{sec:resultsresults}


\begin{figure*}[t]
        \centering
        \begin{minipage}[b]{.65\textwidth}
        {\small
           \begin{tabular}{|l|c|c|c||c|c||c||c|c|}
            \hline
             &\multicolumn{5}{c||}{\textbf{2015}} &\multicolumn{3}{c|}{\textbf{2017}}\\
            \textbf{Metric}       & W'15 & F'16 & D'17 & PD & FTD & vN'17 & PD & FTD \\
            \hline
            Smatch       & 67 & 67 & 64 & \textbf{70} & \textbf{70} & \textbf{71} & \textbf{71} & 70\\
            Unlabeled    & 69 & 69 & 69 & \textbf{73} & \textbf{73} & \textbf{74} & \textbf{74} & \textbf{74} \\
            No WSD       & 64 & 68 & 65 & \textbf{71} & 70 & \textbf{72} & \textbf{72} & 70\\
            Named Ent.   & 75 & 79 & \textbf{83} & 79 & 78 & \textbf{79} & 78 & 77\\
            Wikification &  0 & \textbf{75} & 64 & 71 & 72 & 65 & \textbf{71} & \textbf{71} \\
            Negations    & 18 & 45 & 48 & \textbf{52} & \textbf{52} & \textbf{62} & 57 & 55\\
            Concepts     & 80 & 83 & 83 & 83 & \textbf{84} &82 & \textbf{84} & \textbf{84}\\
            Reentrancies & 41 & 42 & 41 & \textbf{46} & 44 & \textbf{52} & 49 & 46\\
            SRL          & 60 & 60 & 56 & \textbf{63} & 61 & \textbf{66} & 64 & 62\\
            \hline
        \end{tabular}
        }
        \captionof{table}{Details for the LDC2015E86 and LDC2017T10 test sets}\label{table:damonte}
        \end{minipage}\hfill
    \begin{minipage}[b]{.3\textwidth}
        \includegraphics[scale=0.4]{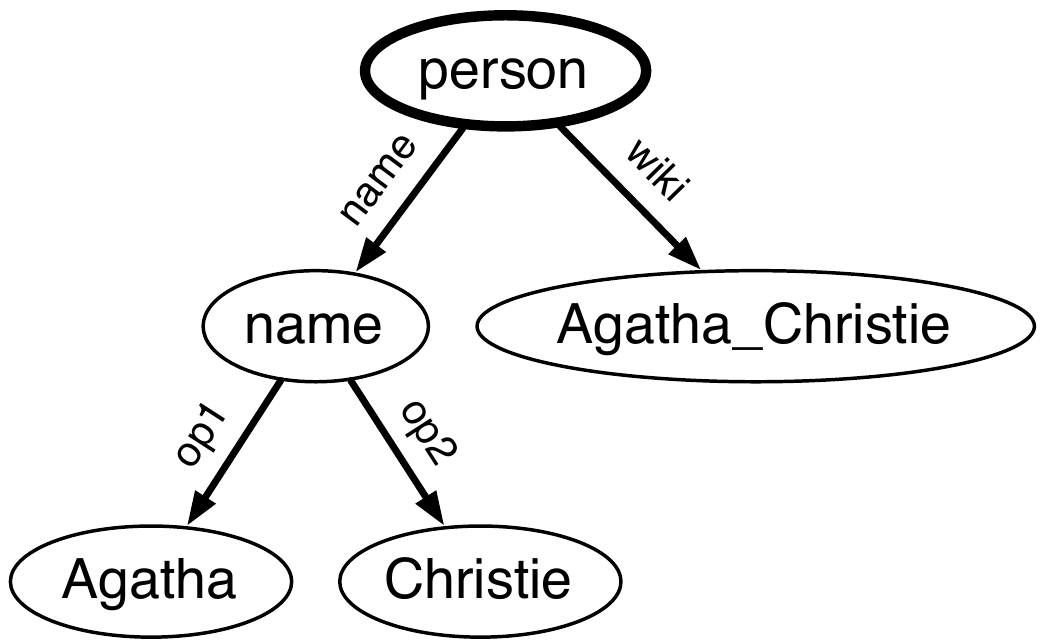}
\caption{A named entity}\label{fig:namedEnt}
	\end{minipage}
    
    \vspace{-1em}

\end{figure*}

Table~\ref{table:smatch} shows the Smatch scores \citep{CaiK13} of our
models, compared to a selection of previously published results. Our results are averages over 4 runs with $95\%$ confidence intervals (JAMR-style baselines are single runs). On the 2015 dataset, our
best models (local + projective, \kg\ + fixed-tree) outperform all
previous work, with the exception of the \citet{foland2017abstract}
model; on the 2017 set we match state of the art results (though note that \newcite{van2017neural} use 100k additional sentences of silver data).
The fixed-tree decoder seems to work well with either edge model, but performance of the projective decoder drops with the \kg\ edge scores. It may be that, while the hinge loss used in the \kg\ edge
scoring model is useful to finding the correct unlabeled dependency
tree in the fixed-tree decoder, scores for bad edges -- which are
never used when computing
the hinge loss -- are not trained
accurately. Thus such edges may be erroneously used by the
projective decoder.

As expected, the type-unaware baseline has low recall, due to its
inability to produce well-typed trees. The fact that our models
outperform the JAMR-style baseline so clearly is an indication that
they indeed gain some of their accuracy from the type information in
the elementary as-graphs, confirming our hypothesis that an explicit
model of the compositional structure of the AMR can help the parser
learn an accurate model.

Table~\ref{table:damonte} analyzes the performance of our two best
systems (PD = projective, FTD = fixed-tree) in more detail, using the
categories of \newcite{E17-1051}, and compares them to Wang's,
Flanigan's, and Damonte's AMR parsers on the 2015 set and , and \newcite{van2017neural} for the 2017 dataset. (\newcite{foland2017abstract} did not publish such results.) The good scores we
achieve on reentrancy identification, despite removing a large amount of  reentrant edges from the training data, indicates that our elementary
as-graphs successfully encode phenomena such as control and
coordination.

The projective decoder is given 4, and the fixed-tree decoder 6,
supertags for each token. We trained the supertagging and edge
scoring models of Section~\ref{sec:training} separately; joint training did not help. Not sampling the supertag types $\tau_i$ during training of the \kg\ model, removing them from
the input, and removing the character-based LSTM encodings
$c_i$ from the input of the supertagger, all reduced our models' accuracy.

\subsection{Differences between the parsers}

Although the Smatch scores for our two best models are close, they
sometimes struggle with different sentences. The fixed-tree parser is
at the mercy of the fixed tree; the projective parser cannot produce
non-projective AM dependency trees. It is remarkable that the
projective parser does so well, given the prevalence of non-projective
trees in the training data. Looking at its analyses, we find that it
frequently manages to find a projective tree which yields an (almost)
correct AMR, by choosing supertags with unusual types, and by using
modify rather than apply (or vice versa).




\section{Conclusion} \label{sec:conclusion}

We presented an AMR parser which applies methods from supertagging and
dependency parsing to map a string into a well-typed AM term, which it
then evaluates into an AMR. The AM term represents the compositional
semantic structure of the AMR explicitly, allowing us to use standard
tree-based parsing techniques.

The projective parser currently computes the complete parse chart. In
future work, we will speed it up through the use of pruning
techniques. We will also look into more principled methods for
splitting the AMRs into elementary as-graphs to replace our
hand-crafted heuristics. In particular, advanced methods for alignments, as in \newcite{lyu2018amr}, seem promising. Overcoming the need for heuristics also seems to be a crucial ingredient for applying our method to other semantic representations.


\paragraph{Acknowledgements} \label{sec:acknowledgements}
We would like to thank the anonymous reviewers for their comments. We thank Stefan Gr\"unewald for his contribution to our PyTorch implementation, and want to acknowledge the inspiration obtained from \newcite{nguyen2017novel}. We also extend our thanks to the organizers and participants of the Oslo CAS Meaning Construction workshop on Universal Dependencies.
This work was supported by the DFG grant KO 2916/2-1 and a Macquarie University Research Excellence Scholarship for Jonas Groschwitz.

\bibliographystyle{acl}
\bibliography{mybib}

\clearpage

\appendices
\FloatBarrier

\section{NP-completeness of the decoding problem}
\label{sec:npc}

We prove NP-completeness for the well-typed decoding problem by
reduction from \textsc{Hamiltonian-Path}.

Let $G = (V,E)$ be a directed graph with nodes $V = \{1,\ldots,n\}$
and edges $E \subseteq V \times V$. A Hamiltonian path in $G$ is a
sequence $(v_1,\ldots,v_n)$ that contains each node of $V$ exactly
once, such that $(v_i,v_{i+1}) \in E$ for all $1 \leq i \leq n-1$. We
assume w.l.o.g.\ that $v_n=n$. Deciding whether $G$ has a Hamiltonian
path is NP-complete.

\begin{figure}
	\centering
	\includegraphics[width=\columnwidth]{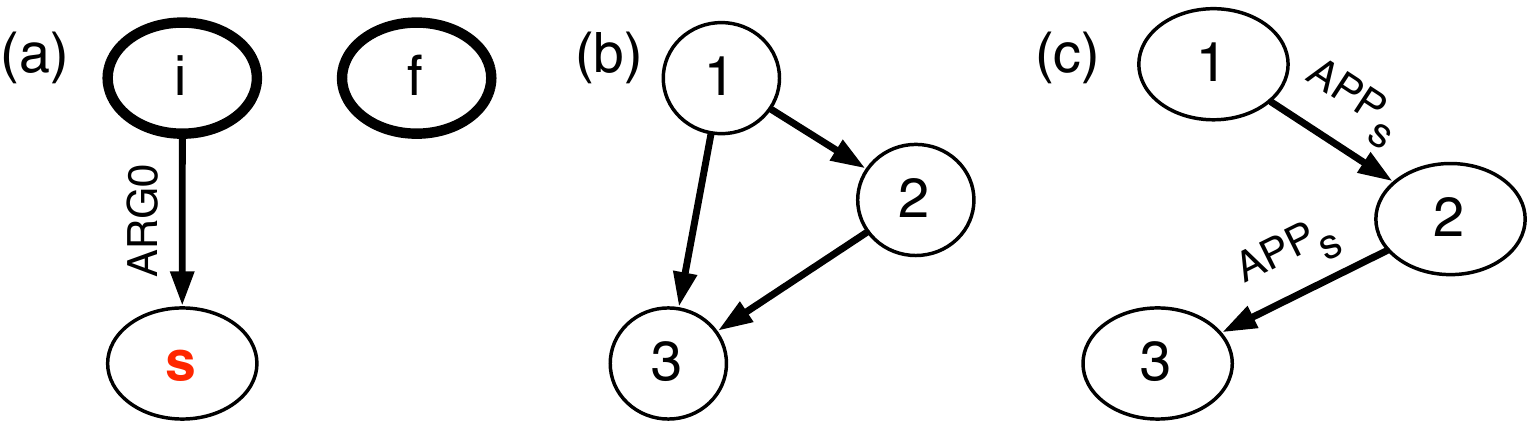}
	\caption{(a) The two graph fragments required for the
		NP-completeness proof. (b) An example graph and (c) the AM
		dependency tree corresponding to its Hamiltonian path.}
	\label{fig:npc-graphs}
\end{figure}

Given $G$, we construct an instance of the decoding problem for the
sentence $w = 1 \ldots n$ as follows. We assume that the first graph
fragments shown in Fig.~\ref{fig:npc-graphs}a (with node label ``i'')
is the only graph fragment the supertagger allows for 1, \ldots,
$n-1$, and the second one (with node label ``f'') is the only graph
fragment allowed for $n$. We let $\escore{i}{k} = 1$ if
$(i,k) \in E$, and zero otherwise.

Under this construction, every well-typed AM dependency tree for $w$
corresponds to a linear sequence of nodes connected by edges with
label $\app{s}$ (see Fig.~\ref{fig:npc-graphs}c for an example) More
specifically, $n$ is a leaf, and every node except for $n$ has
precisely one outgoing $\app{s}$ edge; this is enforced by the
well-typedness. Because of the edge scores, the score of such a
dependency tree is $n-1$ iff it only uses edges that also exist in
$G$; otherwise the score is less than $n-1$. Therefore, we can decide
whether $G$ has a Hamiltonian path by running the decoder,
i.e. computing the highest-scoring well-typed AM dependency tree $t$
for $w$, and checking whether the score of $t$ is $n-1$.

\section{Neural Network Details} \label{sec:hyperparameters}

We implemented the supertagger (Section 5.1) and the local dependency model (Section 5.3) in PyTorch, and used the original DyNet implementation of \cite{kiperwasser16:_simpl_accur_depen_parsin_using} (short \kg) for the \kg\ model.
Further details are:

\begin{enumerate}
	\item As pre-trained embeddings, we use GloVE \cite{pennington2014glove}. The vectors we use have 200 dimensions and are trained on Wikipedia and Gigaword. We add randomly initialized vectors for the \tkn{name}, \tkn{date} and \tkn{number} tokens and for the unknown word token (if no GloVE vector exists). We keep these embeddings fixed and do not train them.
	\item For the learned word embeddings, we follow \kg\ in all our models in using a word dropout of $\alpha=0.25$. That is, during training, for a word that occurs $k$ times in the training data, with probability $\frac{\alpha}{k+\alpha}$ we instead use the word embedding for the unknown word token instead of $w_i$.
	\item The character-based encodings $c_i$ for the supertagger are generated by a single layer LSTM with 100 hidden dimensions, reading the word left to right. If a word (or sequence of words) is replaced by e.g. a \tkn{name} token during pre-processing, the character-based encoding reads the original string instead (this helps to classify names correctly as country, person etc.).
	\item To prevent overfitting, we add dropout of 0.5 in the LSTM layers of all the models except for the \kg\ model which we keep as implemented by the authors. We also add 0.5 dropout to the MLPs in the supertagger and local dependency model.
    \item For the \kg\ model with the fixed-tree decoder, we perform early stopping computing the Smatch score on the development set with 2 best supertags after each epoch.
	\item Hyperparameters for the different neural models are detailed in Tables \ref{tbl:supertagger}, \ref{tbl:kg} and \ref{tbl:simplified}. We did not observe any improvements when increasing the number of LSTM dimensions of the \kg\ model.
\end{enumerate}

\begin{table}
\begin{tabular}{|l|l|}
\hline
Optimizer & Adam \\
Learning Rate & 0.004 \\
Epochs & 37 \\
Pre-trained word embeddings & glove.6B \\
Pre-trained word emb. dimension & 200 \\
Learned word emb. dimension & 100 \\
POS embedding dimension & 32 \\
Character encoding dimension & 100 \\
$\alpha$ (word dropout) & 0.25 \\
Bi-LSTM layers & 2 (stacked) \\
Hidden dimensions in each LSTM & 256 \\
Hidden units in MLPs & 256 \\
Internal dropout of LSTMs, MLPs & 0.5 \\
Input vector dropout & 0.8 \\
\hline
\end{tabular}
\caption{Hyperparameters used for training the supertagger (Section 5.1)}\label{tbl:supertagger}
\end{table}

\begin{table}
	\begin{tabular}{|l|l|}
		\hline
		Optimizer & Adam \\
		Learning rate & default \\
		Epochs & 16 \\
		Word embedding dimension & 100 \\
		POS embedding dimension & 20 \\
		Type embedding dimension & 32 \\
		$\alpha$ (word dropout) & 0.25 \\
		Bi-LSTM layers & 2 (stacked) \\
		Hidden dimensions in each LSTM & 128 \\
		$\delta$ & 0.2 \\
		Hidden units in MLPs & 100 \\
		\hline
	\end{tabular}
	\caption{Hyperparameters used for training \kg\'s model (Section 5.2)}\label{tbl:kg}
\end{table}

\begin{table}
\begin{tabular}{|l|l|}
\hline
Optimizer & Adam \\
Learning Rate & 0.004 \\
Epochs & 35 \\
Pre-trained word embeddings & glove.6B \\
Pre-trained word emb. dimension & 200 \\
POS embedding dimension & 25 \\
Bi-LSTM layers & 2 (stacked) \\
Hidden dimensions in each LSTM & 256 \\
Hidden units in MLPs & 256 \\
Internal dropout of LSTMs, MLPs & 0.5 \\
Input vector dropout & 0.8 \\
\hline
\end{tabular}
\caption{Hyperparameters used for training the simplified dependency model (Section 5.3)}\label{tbl:simplified}
\end{table}

\section{Decoding Details}

The goal item of the decoders is one with empty type that covers the complete sentence. In practice, the projective decoder always found such a derivation. However, in in a few cases, this cannot be achieved by the fixed-tree decoder with the given supertags. Thus, we take instead the item which minimizes the number of open sources in the resulting graph. 

When the fixed-tree decoder takes longer than 20 minutes using $k$ best supertags, it is re-run with $k-1$ best supertags. If $k=0$, a dummy graph is used instead. Typically, the limit of 20 minutes is exceeded one or more times by the same sentence of the test set.

With the projective decoder, in most runs, 1 or 2 sentences took too long to parse and we used a dummy graph instead.

We trained the supertagger and all models 4 times with different initializations. For evaluation, we paired each edge model with a supertag model such that every run used a different edge model and different supertags. The reported confidence intervals are 95\% confidence intervals according to the t-distribution.

\section{Pre- and postprocessing Details} \label{sec:pre-post}




\subsection{Aligner}\label{appendix:aligner}
We use a heuristic process to generate alignments satisfying the conditions above. Its core principles are similar to the JAMR aligner of \cite{FlaniganTCDS14}. There are two types of actions:

Action 1: Align a word to a node (based on the word and the node label, using lexical similarity, handwritten rules\footnote{E.g. the node label \nodelabel{have-condition-91} can be aligned to \word{if} and \word{otherwise}.} and WordNet neighbourhood; we align some name and date patterns directly). That node becomes the lexical node of the alignment. 

Action 2: Extend an existing alignment to an adjacent node, such as from \nodelabel{write} to \nodelabel{person} in the example graph in the main paper. Such an extension is chosen on a heuristic based on
\begin{enumerate}
	\item the direction and label of the edge along which the alignment is split,
	\item the labels of both the node we spread from, and the node we spread to, and
	\item the word of the alignment.
\end{enumerate}
We disallow this action if the resulting alignment would violate the single-root constraint of Section 4.2 in the main paper.

Each action has a basic heuristic score, which we increase if a nearby node is already aligned to a nearby word, and decrease if other potential operations conflict with this one. We remove We iteratively execute the highest scoring action until all heuristic options are exhausted or all nodes aligned. 
We then align remaining unaligned nodes to words near adjacent alignments.

\subsection{Postprocessing}\label{appendix:post}
Having obtained an AM dependency tree, we can recover an AM term and evaluate it. During postprocessing we have to re-lexicalize the resulting graph according to the input string. For relatively frequent words in the training data (occurring at least 10 times), we take the supertagger's prediction for the label.
For rarer words, the neural label prediction accuracy drops, and we
simply take the node label observed most often with the word in the
training data. For unseen words, if the lexicalized node has outgoing ARGx edges, we first try to find a verb lemma for the word in WordNet \cite{miller1995wordnet} (we use version 3.0). If that fails, we try, again in WordNet, to find the closest verb derivationally related to any lemma of the word. If that also fails, we take the word literally. In any case, we add \nodelabel{-01} to the label.
If the lexicalized node does not have an outgoing ARGx edge, we try fo find a noun lemma for the word in Wordnet, and otherwise take the word literally.

For names, we again simply look up name nodes and wiki entries
observed for the word in the training data, and for unseen names use
the literal tokens as the name and no wiki entry. We recover dates and
numbers straightforwardly.



\end{document}